\pdfoutput=1

\documentclass[11pt]{article}

\usepackage{acl}

\usepackage{times}
\usepackage{latexsym}
\usepackage{hyperref}
\usepackage{color, colortbl}
\definecolor{Gray}{gray}{0.9}
\definecolor{high}{HTML}{c2d9d6}
\definecolor{low}{HTML}{f9f0c8}
\definecolor{total}{HTML}{dbeda5}
\definecolor{hard}{HTML}{ffebad}
\usepackage{array}
\usepackage{booktabs}
\usepackage{multirow}
\usepackage{amsmath}
\usepackage{amssymb}
\usepackage{graphicx}
\usepackage{enumerate}
\usepackage{diagbox}
\usepackage{mathtools}
\usepackage{paralist}
\usepackage{tabularx}
\usepackage{lipsum}
\usepackage{ragged2e}
\usepackage{makecell}
\usepackage{siunitx}

\usepackage[T1]{fontenc}

\usepackage[utf8]{inputenc}

\usepackage{microtype}

\usepackage{inconsolata}

\usepackage{graphicx}

\newcommand{\dataset}{\textsc{ComparisonQA}}

\title{\dataset: Evaluating Factuality Robustness of LLMs Through Knowledge Frequency Control and Uncertainty}

\author{
  \textbf{Qing Zong, \qquad Zhaowei Wang, \qquad Tianshi Zheng,} \\ \textbf{Xiyu Ren, \qquad Yangqiu Song} \\
  Department of Computer Science and Engineering, HKUST\\ 
  \texttt{\{qzong, yqsong\}@cse.ust.hk}
  }

\begin{document}
\maketitle
\begin{abstract}

The rapid development of LLMs has sparked extensive research into their factual knowledge. Current works find that LLMs fall short on questions around low-frequency entities. However, such proofs are unreliable since the questions can differ not only in entity frequency but also in difficulty themselves. So we introduce \textbf{\dataset{}} benchmark, containing \textbf{283K} abstract questions, each instantiated by a pair of high-frequency and low-frequency entities. It ensures a controllable comparison to study the role of knowledge frequency in the performance of LLMs. Because the difference between such a pair is only the entity with different frequencies. 
In addition, we use both correctness and uncertainty to develop a two-round method to evaluate LLMs' knowledge robustness. It aims to avoid possible semantic shortcuts which is a serious problem of current QA study. Experiments reveal that LLMs, including GPT-4o, exhibit particularly low robustness regarding low-frequency knowledge. 
Besides, we find that uncertainty can be used to effectively identify high-quality and shortcut-free questions while maintaining the data size. Based on this, we propose an automatic method to select such questions to form a subset called \textbf{\dataset-Hard}, containing only hard low-frequency questions.
\footnote{\url{https://github.com/HKUST-KnowComp/ComparisonQA}} 

\end{abstract}

\section{Introduction}

The rapid advancement of large language models (LLMs) has promoted a lot of study on their factual knowledge and reasoning ability~\cite{SimpleQA,LongFact,MMLU}. 

\begin{figure}[t]
\centering
\includegraphics[scale=0.35]{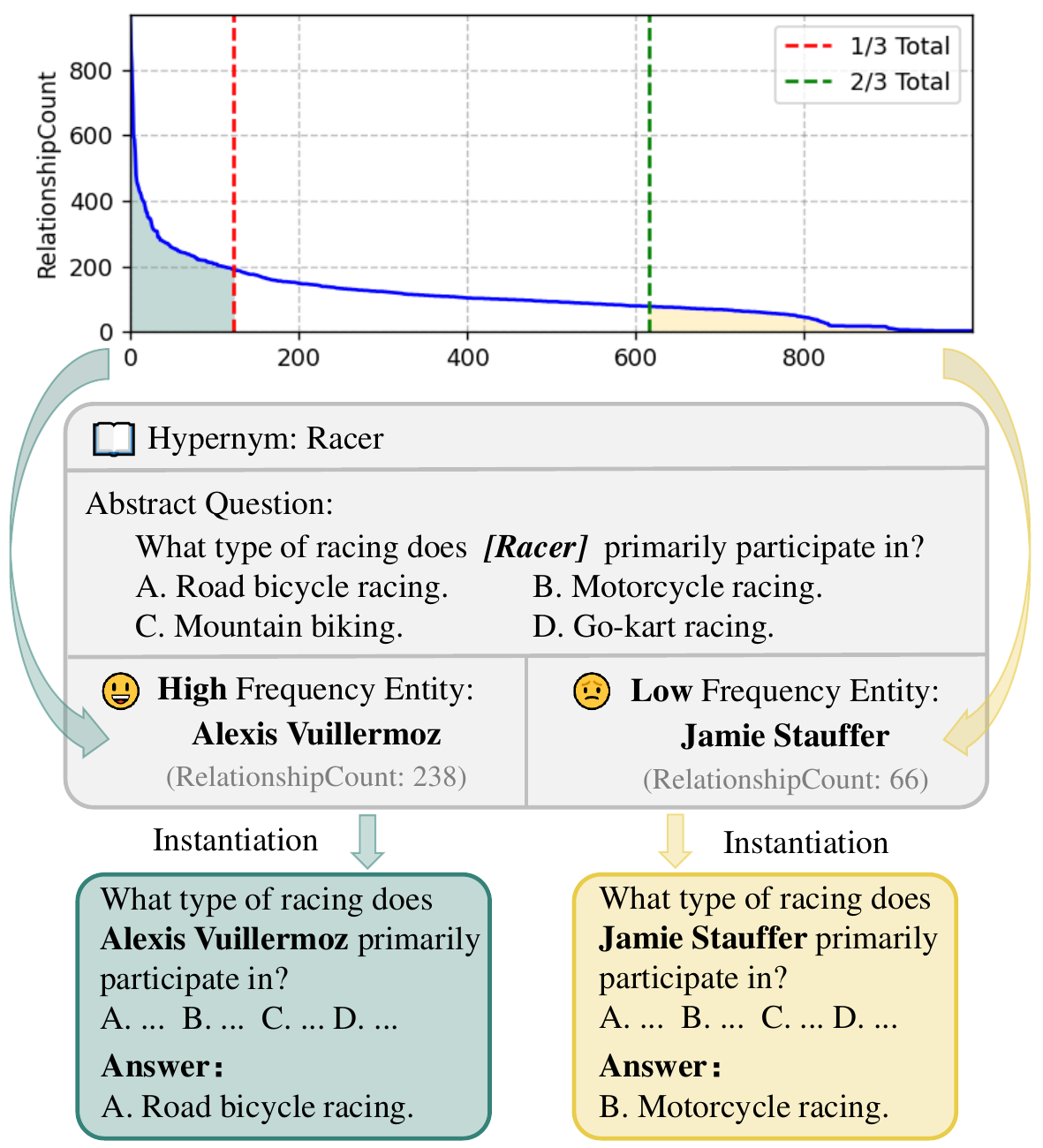}
\caption{An example from \dataset{}} 
\label{dataset_image}
\vspace{-0.2in}
\end{figure} 

\citet{headtotail} and \citet{DBLP:conf/acl/MallenAZDKH23} compare LLMs' performance on questions around entities with different frequencies. 
LLMs are found struggle to handle tail knowledge.
However, the questions they study are different and can vary in difficulty levels, not only in entity frequencies. 
As \citet{allenzhu} emphasized, we need a more ``\textit{controlled, synthetic experiment that confirms the weakness of LLMs}'' nowadays.  
Therefore, existing comparisons are not enough since they can not guarantee that the low frequency is the only cause of LLMs' poor performance.

To tackle the issue, we introduce our \textbf{\dataset{}} benchmark. Pairs of high-frequency and low-frequency entities share the same question, as shown in Figure~\ref{dataset_image}. 
The shared abstract question, with a hypernym to represent the two specific entities, guarantees that the difference between such a pair is only the entity. This allows for a controllable comparison between high-frequency and low-frequency entities. \dataset{} is a large scale dataset containing \textbf{283K} such question pairs. 
It is generated through an automatic pipeline base on raw knowledge base, ensuring both diversity and scalability. Through this benchmark, we can completely compare LLM's performance on different knowledge frequencies.

For a more robust and accurate evaluation, we use the multiple-choice format~\cite{MMLU, strategyqa} in our benchmark. But semantic shortcuts~\cite{shortcutlearning} between questions and options may help LLMs to guess the answer, which is also a common but severe problem recently. 
Thus, we further design a two-round method using both correctness and uncertainty to evaluate LLMs' knowledge. During experiments, we found that LLMs have very poor robustness, especially on low-frequency knowledge, where even the powerful GPT-4o also performs badly.

Recent benchmarks, like SimpleQA~\cite{SimpleQA}, ensure their difficulty by collecting questions adversarially against LLMs’ responses. But relying only on accuracy, they ignore the quality of their questions, and the difficulty is closely related to the models they use. 
Fortunately, our experiments find that uncertainty is also an effective tool in selecting both high-quality and shortcut-free questions while maintaining the benchmark size. 
Combining accuracy and uncertainty, we propose a new flexible method to select our subset called \textbf{\dataset-Hard} for future study. It contains \textbf{81K} difficult low-frequency questions with high-quality and no semantic shortcuts. 

In summary, we have three main contributions:

\noindent\textbf{(1) [Resource]} We introduce \dataset{} benchmark, where a pair of entities share the same abstract question. It enables a more controllable and reasonable proof that LLMs perform worse when the required knowledge is less frequent. (\S\ref{ComparisonQA})

\noindent\textbf{(2) [Method]} We design a two-round method using correctness and uncertainty to evaluate LLMs' robust knowledge. \textbf{[Finding]} LLMs can not stand such a test, especially on low-frequency knowledge, where even GPT-4o performs badly. (\S\ref{sec: Robust Knowledge Measurement})

\noindent\textbf{(3) [Finding]} Uncertainty is more helpful to find questions with high quality. \noindent\textbf{[Resource]} Through this, we select \dataset{}-Hard benchmark containing only hard and low-frequency questions of high quality and no shortcuts. (\S\ref{sec:ComparisonQA_Hard})


\section{Related Works}
\subsection{Benchmarking LLMs' Factuality}
The factuality evaluation of LLMs has recently attracted significant attention.
 Some factuality benchmarks require open-ended generation by LLMs, such as SimpleQA~\cite{SimpleQA}, 
 SelfAware~\cite{selfaware}, CLR-Fact~\cite{clrfact}, and HaluEval~\cite{HaluEval}. Such evaluations either rely heavily on expert annotation, or utilize automatic answer matching that sacrifices evaluation accuracy~\cite{factsocre,factool,Factcheck-GPT}. Other benchmarks adopt the format of Yes-or-No questions~\cite{strategyqa,snowball} or multiple-choice questions (MCQ)~\cite{MMLU,mmlupro,DBLP:conf/iclr/Zheng0M0H24}. These formats allow model responses to be easily parsed and compared with gold labels, enabling solid yet efficient evaluations.


\begin{figure*}[t]
\centering
\includegraphics[scale=0.55]{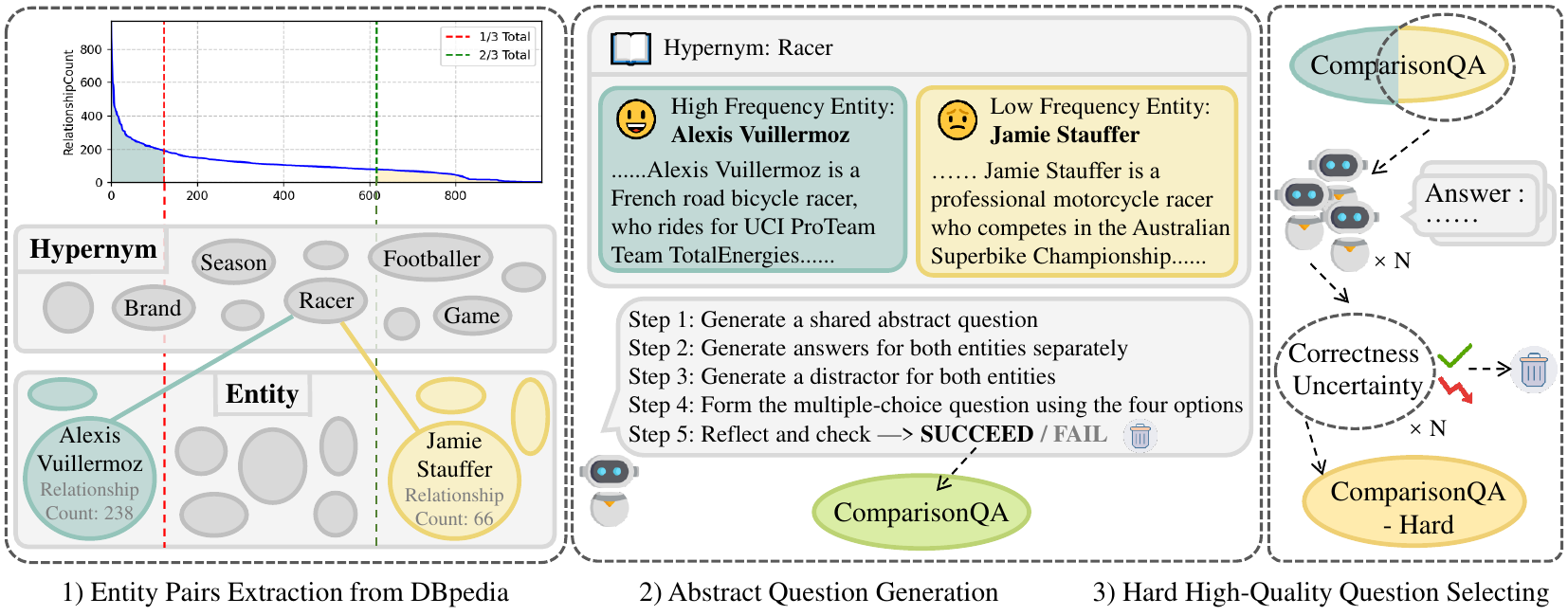} 
\caption{An overview of our benchmark curation pipeline. It contains three parts. Through the first two parts, (1) Entity Pairs Extraction from DBpedia and (2) Abstract Question Generation, we can get the whole \dataset{}. And through the third part, (3) Hard High-Quality Question Selecting, we can get a harder subset, containing only difficult low-frequency questions with high quality and no semantic shortcut.} 
\label{pipeline}
\end{figure*}

\subsection{Long-Tail Knowledge}

Long-tail knowledge~\cite{LongFact,DBLP:journals/corr/abs-2306-17472} is an important aspect of factuality. \citet{longtailqaG} proposes an automatic approach to generate questions for tail entities. \citet{DBLP:conf/icml/KandpalDRWR23} find LLMs struggle to learn long-tail knowledge. Other works study the influence of knowledge frequency:
\citet{DBLP:conf/acl/MallenAZDKH23} introduced PopQA, a long-tail benchmark, and found that models' performance will change with the frequency of entities in the questions. \citet{headtotail} also proves this by constructing questions around head, torso, and tail entities.
However, questions in these benchmarks are all in the form of open-ended generation, which can not be easily evaluated.
They also depend on limited number of templates to produce QA questions from knowledge graphs, which will significantly harm the diversity of the benchmarks.
Most importantly, the questions are different, so they may vary in difficulty levels, and thus can not provide fair comparisons.

\subsection{Abstraction Knowledge}

Existing works have studied various aspects of abstraction, for example, entity abstraction~\cite{probase, DBLP:conf/ijcai/SongWW15, DBLP:conf/aaai/XuCH23}, event abstraction~\cite{AbsPyramid,absinstruct}, and conceptual abstraction~\cite{han2024concept}.
Abstraction has been shown to be beneficial for downstream tasks like commonsense reasoning, numerical reasoning, and logical reasoning~\cite{DBLP:journals/corr/abs-2404-00205,DBLP:conf/emnlp/Hong0P0Z24}. 
In this paper, we control question difficulty by sharing the same abstraction form between a pair of entities.

\section{\dataset}
\label{ComparisonQA}

To ensure more fair and controllable comparisons between LLMs' performance on high-frequency and low-frequency factual knowledge, we propose a new benchmark called~\dataset{}. 

\subsection{Question Formulation}
\label{qf}

For an accurate evaluation, our questions are in the form of multiple choice. Although recently there are several generative QA benchmarks, like SimpleQA~\cite{SimpleQA}, and also some automatic methods~\cite{llmasjudge} to evaluate generated answers. There are still significant limitations in such generative QAs since the answer should be single and indisputable. Questions like \textit{What is the primary focus or intention behind Civil Procedure Rules} cannot be included since there are many ways to answer this question. However, multiple-choice questions do not have such limitations.

Each piece of data in~\dataset{}, shown in Figure~\ref{dataset_image}, contains an abstract question shared by two entities having the same hypernym. One entity, having many relationships in DBpedia~\cite{dbpedia}, which will be introduced next, is the high-frequency entity, and the other, with only a few relationships, is the low-frequency one. The question will have different answers for the two entities, respectively.
Such data form can ensure detailed and controllable comparisons between entities with different frequencies, and that is why we call it \textbf{\dataset{}}. Containing \textbf{283K} such question pairs, the benchmark is constructed through a fully automated pipeline, which is cheap and scalable.

\subsection{Curation Pipeline}
In this section, we discuss the data curation pipeline for our dataset.
 As shown in Figure~\ref{pipeline}, the pipeline contains three parts: (1) Entity Pairs Extraction from DBpedia, (2) Abstract Question Generation, and (3) Hard High-Quality Question Filtering. Here we will introduce the first two parts, used to build \dataset{}, and leave the third part, used to build \dataset-Hard, for \S\ref{sec:ComparisonQA_Hard}.

\subsubsection{Entity Pairs Extraction from DBpedia}
Following \citet{headtotail}, an entity's frequency is defined by its number of relationships in DBpedia. 
High-frequency entities are those whose cumulative relationships account for the first $1/3$ of all sorted entities, and low-frequency entities are for the last $1/3$. Details are shown in Appendix~\ref{sec:freq}.

In practice, we first get all the hypernyms in DBpedia and classify the entities belonging to them into high-frequent and low-frequent. 
Then, we map the entities one-by-one to get entity pairs. 
The reason behind this is that not every random pair of entities can have a shared abstract question easily. 
For example, it's hard to write a shared question for \textit{Einstein} and \textit{Apple} even though they are all high-frequency entities. 
But it's easy for two specific universities to share a same abstract question.
In order for our entity pairs to produce high quality abstract questions, we ensure that the two entities have the same hypernym. 
This allows them to have similar descriptions, making it easier to generate an abstract question.

\subsubsection{Abstract Question Generation}
\label{sec:Abstract Question Generation}

After acquiring the entity pairs, we adopt a multi-step curation pipeline to generate high-quality abstract multiple-choice questions: (1) Generate an abstract question~(without options) according to the DBpedia descriptions of both entity. (2) Separately generate the corresponding answers based on the descriptions, with length control to alleviate bias among candidate answers. (3) Generate distractors for both entities with length control. Compared to randomly selected distractors, LLM-selected distractors generation have demonstrated its effectiveness for the high relevance between distractors and designated choices \cite{zheng2024assessingrobustnessretrievalaugmentedgeneration}. (4) Formulate the final multiple-choice question using the four answer candidates above. (5) Proofread the question according to the standards below.

The standards for questions are presented as follows: (1) Quality: The questions should have one and only one correct answer for both entities. (2) Semantic Shortcuts: The correct answer cannot be simply guessed by the names of the entities or the way the questions are asked. For example, the question: \textit{What was the primary operational location of Sydney O-Class Tram?  A. Oslo, Norway B. Sydney, Australia C. Stockholm, Sweden D. Melbourne, Australia} is not allowed since only the correct answer contains \textit{Sydney} which is also in the entity's name. (3) Length Bias: The four options in one question should have roughly the same length to avoid length bias. In summary, the generated questions should have high-quality and no shortcuts, in semantics or length.  

 During the curation, we utilize \textit{GPT-4o-mini} \cite{GPT4omini} with few-shot expert-written Chain-of-thought (CoT) demonstrations \cite{COT}, with details provided in Appendix~\ref{sec:generatin}.
 
\subsection{Expert Verification}
\label{sec:Expert Verification}

We enlist the help of three postgraduate students, each with extensive experience in NLP research, to validate the quality of these generated questions through a sample of 200 question pairs. The instruction is the same as the \textit{standard (1)} given to LLMs above.
The quality of each pair is decided by majority voting. 
Results show that their total agreement~(all 3 experts have the same judgement) is 88.0\%. And 95.5\% of the abstract questions are considered correct and of high quality both for the high-frequency entity and the low-frequency entity, demonstrating the reliability of our benchmark.


\subsection{Main Evaluations}

\dataset{} is a large-scale benchmark comprising a total of 283,455 abstract questions, each paired with a high-frequency and a low-frequency instantiation. 
Detailed statistics are in Appendix~\ref{sec:statistics}.

We experiment with a selection of LLMs on our \dataset{} benchmark to investigate their performance on high-frequency and low-frequency questions and also the difference between them.

\begin{table*}[t]
    \huge
    \centering
    \resizebox{\linewidth}{!}{
	\begin{tabular}{@{}l||c>{\columncolor{high}}ccc>{\columncolor{low}}cc|c>{\columncolor{total}}cc||ccc@{}}
	\toprule
    \multirow{3}{*}{\textbf{Models}}&\multicolumn{3}{c}{\textbf{High Freq Question}} &\multicolumn{3}{c|}{\textbf{Low Freq Question}}&\multicolumn{3}{c|}{\textbf{Average}}&\multicolumn{3}{c}{\textbf{Difference (H --> T)}}
    \\
	&\textbf{Unc.}&\cellcolor{white}\textbf{Acc}&\textbf{Ma-F1}&\textbf{Unc.}&\cellcolor{white}\textbf{Acc}&\textbf{Ma-F1}&\textbf{Unc.}&\cellcolor{white}\textbf{Acc}&\textbf{Ma-F1}&\textbf{Unc.}&\cellcolor{white}\textbf{Acc}&\textbf{Ma-F1}\\
    &\textbf{(↓)}&\cellcolor{white}\textbf{(↑)}&\textbf{(↑)}&\textbf{(↓)}&\cellcolor{white}\textbf{(↑)}&\textbf{(↑)}&\textbf{(↓)}&\cellcolor{white}\textbf{(↑)}&\textbf{(↑)}&&&\\
            \midrule
            Random & - & 25.29 & 25.29 & - & 25.22 & 25.22 & - & 25.26 & 25.26 & - & ↓ 0.07 & ↓ 0.07 \\
            Majority & - & 25.70 & 10.22 & - & 25.14 & 10.04 & - & 25.42 & 10.13 & - & ↓ 0.56 & ↓ 0.18 \\
		  \bottomrule
          \rowcolor[gray]{0.9} \multicolumn{13}{c}{\textbf{LLM (Open Source) + Zero-Shot}} \\
          \toprule
            Llama-3 \Large{\textit{8B}} & 54.33 & 65.90 & 63.60 & 81.54 & 53.83 & 51.29 & 67.94 & 59.87  & 57.44  & ↑ \phantom{0}6.11 & ↓ 12.07 & ↓ 12.30 \\
            Llama-3-Instruct \Large{\textit{8B}} & 77.55 & \underline{80.72} & \underline{80.71} & 117.41 & 69.03 & 68.95 & 97.48  & 74.88 & 74.83 & ↑ 39.86 & ↓ 11.69 & ↓ 11.76 \\
            Llama-3.1 \Large{\textit{8B}} & 55.91 & 65.29 & 63.31 & 83.35 & 52.66 & 50.52 & 69.63  & 58.98  & 56.92  & ↑ 27.44 & ↓ 12.63 & ↓ 12.78 \\
            Llama-3.1-Instruct \Large{\textit{8B}} & 58.97 & 80.06 & 80.08 & 87.05 & \underline{69.99} & \underline{69.94} & 73.01  & \underline{75.03}  & \underline{75.01}  & ↑ 28.08 & ↓ 10.07 & ↓ 10.14 \\
            Gemma-2 \Large{\textit{9B}} & 124.97 & 64.50 & 64.80 & 211.34 & 52.24 & 51.23 & 168.16  & 58.37  & 58.01  & ↑ \underline{\textbf{86.37}} & ↓ 12.26 & ↓ \underline{13.57} \\
            Phi-3.5-mini-Instruct \Large{\textit{4B}} & \underline{27.81} & 72.81 & 72.78 & \underline{39.80} & 65.26 & 65.00 & \underline{33.81}  & 69.04  & 68.89  & ↑ 11.99 & ↓ \phantom{0}7.55 & ↓ \phantom{0}7.78 \\
            Falcon2 \Large{\textit{11B}} & 56.93 & 70.72 & 69.80 & 87.31 & 58.07 & 56.70 & 72.12  & 64.40  &  63.25 & ↑ 30.38 & ↓ \underline{12.65} & ↓ 13.10 \\
            Mistral-v0.3 \Large{\textit{7B}} & 39.83 & 65.55 & 63.11 & 56.99 & 53.36 & 50.26 & 48.41  & 59.46  & 56.69  & ↑ 17.16 & ↓ 12.19 & ↓ 12.85 \\
            Mistral-v0.3-Instruct \Large{\textit{7B}} & 44.73 & 73.53 & 73.14 & 66.09 & 63.05 & 62.40 & 55.41  & 68.29  & 67.77  & ↑ 21.36 & ↓ 10.48 & ↓ 10.74\\
            \bottomrule
          \rowcolor[gray]{0.9} \multicolumn{13}{c}{\textbf{LLM (Open Source) + Few-Shot}} \\
          \toprule
            Llama-3 \Large{\textit{8B}} & 21.89 & 75.57 & 75.55 & 23.75 & 61.00 & 61.01 & 22.82  & 68.29  & 68.28  & ↑ 1.86 & ↓ \underline{\textbf{14.57}} & ↓ \underline{\textbf{14.55}} \\
            Llama-3-Instruct \Large{\textit{8B}} & 26.20 & 79.94 & 79.92 & 28.70 & 67.98 & 67.95 & 27.45  & 73.96  & 73.93  & ↑ \underline{2.50} & ↓ 11.96 & ↓ 11.96 \\
            Llama-3.1 \Large{\textit{8B}} & 20.82 & 74.91 & 74.89 & 22.62 & 62.00 & 62.00 & 21.72  & 68.46  & 68.45  & ↑ 1.81 & ↓ 12.91 & ↓ 12.90 \\
            Llama-3.1-Instruct \Large{\textit{8B}} & 20.63 & 79.74 & 79.74 & 22.48 & \underline{69.09} & \underline{69.07} & 21.56 & \underline{74.42}  & \underline{74.40}  & ↑ 1.85 & ↓ 10.65 & ↓ 10.66 \\
            Gemma-2 \Large{\textit{9B}} & 20.26 & \underline{80.10} & \underline{80.08} & 22.36 & 68.36 & 68.31 & 21.31  & 74.23  & 74.20  & ↑ 2.10 & ↓ 11.74 & ↓ 11.77 \\
            Phi-3.5-mini-Instruct \Large{\textit{4B}} & \underline{11.02} & 73.68 & 73.67 & \underline{\textbf{11.85}} & 67.46 & 67.33 & \underline{11.44}  & 70.57  & 70.50  & ↑ 0.83 & ↓ \phantom{0}6.22 & ↓ \phantom{0}6.34 \\
            Falcon2 \Large{\textit{11B}} & 16.42 & 77.11 & 77.01 & 17.77 & 65.92 & 65.75 & 17.10  & 71.52  &  71.38 & ↑ 1.35 & ↓ 11.19 & ↓ 11.26 \\
            Mistral-v0.3 \Large{\textit{7B}} & 13.89 & 75.55 & 75.53 & 14.99 & 62.88 & 62.85 & 14.44  & 69.22  & 69.19   & ↑ 1.10 & ↓ 12.67 & ↓ 12.68\\
            Mistral-v0.3-Instruct \Large{\textit{7B}} & 15.97 & 74.46 & 74.40 & 17.40 & 65.51 & 65.35 & 16.69  & 69.99  & 69.87  & ↑ 1.43 & ↓ \phantom{0}8.95 & ↓ \phantom{0}9.05\\
            \bottomrule
          \rowcolor[gray]{0.9} \multicolumn{13}{c}{\textbf{LLM (Proprietary) API}} \\
          \toprule
            GPT4o-mini (Zero-Shot) & 13.52 & 85.61 & 85.58 & 18.34 & 73.85 & 73.73 & 15.93  & 79.73  & 79.66  & ↑ \phantom{0}4.82 & ↓ 11.76 & ↓ 11.85 \\
            GPT4o-mini (Few-Shot) & 25.74 & 84.78 & 84.69 & 38.17 & 72.76 & 72.47 & 31.96  & 78.77  & 78.58  & ↑ 12.43 & ↓ \underline{12.02} & ↓ \underline{12.22} \\
            GPT4o-mini (CoT) & 10.53 & 86.25 & 86.25 & \underline{12.27} & 74.39 & 74.40 & \underline{\textbf{11.40}} & 80.32 & 80.32 & ↑ \phantom{0}1.74 & ↓ 11.85 & ↓ 11.85\\
            GPT4o (Zero-Shot) & 14.18 & \underline{\textbf{93.86}} & \underline{\textbf{93.95}} & 30.98 & 85.76 & 86.69 & 22.58  & 89.81  & 90.32  & ↑ 16.80 & ↓ \phantom{0}8.10 & ↓ \phantom{0}7.26\\
            GPT4o (Few-Shot) & 28.41 & 93.94 & \underline{\textbf{93.95}} & 45.81 & \underline{\textbf{86.54}} & \underline{\textbf{86.75}} & 37.11  & \underline{\textbf{90.24}}  & \underline{\textbf{90.35}}  & ↑ \underline{17.40} & ↓ \phantom{0}7.40 & ↓ \phantom{0}7.20\\
            GPT4o (CoT) & \underline{\textbf{10.39}} & 92.40 & 92.47 & 18.36 & 85.47 & 85.72 & 14.38 & 88.93 & 89.10 & ↑ \phantom{0}7.97 & ↓ \phantom{0}6.93 & ↓\phantom{0} 6.75\\
		\bottomrule
	\end{tabular}
    }
\vspace{-0.1in}
\caption{Performance of various LLMs on the testing set of~\dataset{}. Unc., Acc, and Ma-F1 denote Uncertainty, Accuracy, and Macro F1-score. 
The Difference column shows how scores change from high-frequency questions to low-frequency questions. The best performances within each method are \underline{underlined}, and the best among all methods are \textbf{bold-faced}. 
And for the Difference column, We \underline{underline} the largest difference within each method and \textbf{bold} the one among all methods. 
More results can be seen in Table~\ref{tab:main_evaluation_results_all}.} 
\label{tab:main_evaluation_results_part}
\vspace{-0.15in}
\end{table*}

\subsubsection{Experiment Setup}

\noindent\textbf{Metric:}
We calculate Uncertainty, Accuracy, and Macro F1-score between model predictions and ground truth labels. 
We apply perplexity-based uncertainty for open source LLMs and verbalized uncertainty for proprietary LLMs due to several reasons, with details explained in Appendix~\ref{sec:unc}.
Thus, we only compare uncertainty between high and low frequency within each setting separately.

\noindent\textbf{Models:}
We experiment with 16 different models, with a full list in Appendix~\ref{sec:full_Experiments}, and categorize the evaluation into three types: (1) Open Source LLM Zero-Shot~\cite{DBLP:conf/emnlp/QinZ0CYY23}. 
(2) Open Source LLM Few-Shot~\cite{FewShot}. 
(3) Proprietary LLM API.

\subsubsection{Results and Analysis}

Evaluation results are reported in Table~\ref{tab:main_evaluation_results_part}. Our observations include:
\textbf{(1) Huge drop in performance from high-freq to low-freq: } All models suffer a performance decrease from high-frequency questions to low-frequency questions in all three settings. For instance, the accuracy and Macro F1-score of Llama-3-8B drop up to about 14 points in the few-shot setting. Proprietary models like GPT-4o are no exception. These all prove that LLM's performance is closely related to the frequency of the knowledge in the corpus.
\textbf{(2) Increased Uncertainty from high-freq to low-freq: } Similarly, the uncertainty of LLMs all increase from high-frequency questions to low-frequency questions. For example, the uncertainty difference of Gemma-2-9B is up to about 86 points when using zero-shot. The uncertainty in the zero-shot setting is generally higher than in the few-shot setting, and the difference is also more pronounced. We think it may be because LLMs find some familiar examples in the few-shot setting, which decreases their uncertainty. In spite of this, the difference is still clear between high-freq and low-freq. These all prove that LLMs not only perform better but also are more confident about high-frequency knowledge.
\textbf{(3) Few-shot helps a lot only for LLMs without instruction-tuning: } Most non-instruction-tuned LLMs show a huge improvement in performance from zero-shot to few-shot, but performance is similar for those instruction-tuned LLMs. This could be because the few-shot examples only teach LLMs how to do multiple-choice questions, while those instruction-tuned ones have already learned.
\textbf{(4) CoT lowers uncertainty but does not awlays aid performance: } It's obvious that after the CoT inference, LLMs are more sure about their answers. However, results show that the average accuracy of GPT-4o even drops after adding CoT, which means CoT can not always help such factual questions.

\begin{table*}[t]
    \Large
    \centering
    \resizebox{\linewidth}{!}{
	\begin{tabular}{@{}l||cc|c|c||cc|c|c@{}}
	\toprule
    \multirow{2}{*}{\textbf{Models}}&\multicolumn{4}{c||}{\textbf{First Round}} &\multicolumn{4}{c}{\textbf{Second Round}}\\
    &\textbf{High}&\textbf{Low}&\textbf{Avg.}&\textbf{Diff}&\textbf{High}&\textbf{Low}&\textbf{Avg.}&\textbf{Diff}\\
            \bottomrule
          \rowcolor[gray]{0.9} \multicolumn{9}{c}{\textbf{Open Source LLM}} \\
          \toprule
            Llama-3 \large{\textit{8B}}   & 75.57 & 61.00 & 68.29 & ↓ \underline{\textbf{14.57}} & 65.11 \large{(-10.47)} &  43.65 \large{(-17.35)} & 54.38 \large{(-13.91)}& ↓ 21.45 \large{(+6.89)\phantom{0}} \\
            Llama-3-Instruct \large{\textit{8B}} & 79.94 & 67.98 & 73.96 & ↓ 11.96 & 68.13 \large{(-11.82)} &  48.73 \large{\underline{(-19.26)}} & 58.43 \large{\underline{(-15.54)}}& ↓ 19.40 \large{(+7.44)\phantom{0}} \\
            Llama-3.1 \large{\textit{8B}}  & 74.91 & 62.00 & 68.46  & ↓ 12.91 & 65.66 \large{(-9.25)\phantom{0}} & 45.21 \large{(-16.78)} & 55.44 \large{(-13.02)} & ↓ 20.45 \large{(+7.53)\phantom{0}} \\
            Llama-3.1-Instruct \large{\textit{8B}} & 79.74 & \underline{69.09} & \underline{74.42} & ↓ 10.65 & \underline{72.23} \large{(-7.51)\phantom{0}} & \underline{55.04} \large{(-14.06)} & \underline{63.64} \large{(-10.78)} & ↓ 17.20 \large{(+6.55)\phantom{0}}\\
            Llama-3.2 \large{\textit{3B}}  & 68.89 & 57.26 & 63.08 & ↓ 11.63 & 58.89 \large{(-10.00)} & 41.11 \large{(-16.15)} & 50.00 \large{(-13.07)} & ↓ 17.78 \large{(+6.15)\phantom{0}} \\
            Llama-3.2-Instruct \large{\textit{3B}} & 71.43 & 62.12 & 66.78 & ↓ \phantom{0}9.32 &  64.07 \large{(-7.37)\phantom{0}}& 46.68 \large{(-15.44)} & 55.37 \large{(-11.40)} & ↓ 17.39 \large{(+8.07)\phantom{0}}\\
            Gemma-2 \large{\textit{2B}}  & 62.99 & 50.93 & 56.96 & ↓ 12.06 & 52.54 \large{(-10.44)} & 35.40 \large{(-15.53)} & 43.97 \large{(-12.99)} & ↓ 17.14 \large{(+5.09)\phantom{0}} \\
            Gemma-2 \large{\textit{9B}}  & \underline{80.10} & 68.36 & 74.23 & ↓ 11.74 & 71.39 \large{(-8.71)\phantom{0}}& 50.48 \large{(-17.88)} & 60.94 \large{(-13.30)} & ↓ \underline{20.92} \large{\underline{(+9.18)}\phantom{0}}\\
            Phi-3.5-mini \large{\textit{4B}}  & 73.68 & 67.46 & 70.57 & ↓ \phantom{0}6.22 & 66.82 \large{(-6.86)\phantom{0}} & 53.51 \large{(-13.95)} & 60.16 \large{(-10.41)} & ↓ 13.31 \large{(+7.08)\phantom{0}}\\
            Falcon2 \large{\textit{11B}}  & 77.11 & 65.92 & 71.52 & ↓ 11.19 & 65.17 \large{\underline{(-11.94)}} & 46.88 \large{(-19.04)} & 56.02 \large{(-15.49)} &  ↓ 18.29 \large{(+7.10)\phantom{0}} \\
            Mistral-v0.3 \large{\textit{7B}}  & 75.55 & 62.88 & 69.22 & ↓ 12.67 & 67.86 \large{(7.69)\phantom{0}} & 49.05 \large{(-13.83)} & 58.46 \large{(-10.76)} & ↓ 18.82 \large{(+6.15)\phantom{0}} \\
            Mistral-v0.3-Instruct \large{\textit{7B}}  & 74.46 & 65.51 & 69.99 & ↓ \phantom{0}8.95 & 67.29 \large{(-7.17)}\phantom{0} & 51.69 \large{(-13.82)} & 59.49 \large{(-10.50)} & ↓ 15.60 \large{(+6.65)\phantom{0}} \\
            \bottomrule
          \rowcolor[gray]{0.9} \multicolumn{9}{c}{\textbf{Proprietary LLM}} \\
          \toprule
            GPT4o-mini  & 84.78 & 72.76 & 78.77 & ↓ \underline{12.02} & 71.00 \large{(-13.78)} & 38.25 \large{\underline{\textbf{(-34.51)}}} & 54.63 \large{\underline{\textbf{(-24.14)}}} & ↓ \underline{\textbf{32.75}} \large{\underline{\textbf{(+20.73)}}} \\
            GPT4o  & \underline{\textbf{93.94}} & \underline{\textbf{86.54}} & \underline{\textbf{90.24}} & ↓ \phantom{0}7.40 & \underline{\textbf{79.85}} \large{\underline{\textbf{(-14.09)}}} & \underline{\textbf{55.94}} \large{(-30.6)\phantom{0}} & \underline{\textbf{67.90}} \large{(-22.34)} & ↓ 23.91 \large{(+16.51)}\\
		\bottomrule
	\end{tabular}
    }
\vspace{-0.07in}
\caption{Accuracy scores in LLMs' robust knowledge measurement. We also report the changes in the scores from the first round to the second round. The best performances within each method are \underline{underlined}, and the best among all methods are \textbf{bold-faced}.  And for the Difference column and values in parentheses, We \underline{underline} the largest difference within each method and \textbf{bold} the one among all methods.} 
\label{tab:robust_knowledge_results}
\vspace{-0.07in}
\end{table*}

\section{Robust Knowledge Measurement}
\label{sec: Robust Knowledge Measurement}

With the help of \dataset{}, we can conduct a more detailed and controllable study of the factual knowledge of LLMs. 

When considering how humans tackle multiple-choice questions, it's often the case that we do not really know the correct answers. Instead, we rely on semantic shortcuts within the questions to make educated guesses. Although we intentionally exclude these shortcuts when constructing the benchmark, they are difficult to eliminate entirely from multiple-choice questions. Such a situation often occurs even in human exam questions. Therefore, we need to devise an effective method to evaluate the robustness of factual knowledge of LLMs in the form of multiple-choice questions.

\subsection{The Definition of Robust Knowledge}
\label{sec:Definition of Robust Knowledge}

We categorize LLM's results into four scenarios based on its uncertainty and the correctness of its answers. The uncertainty pertains to the statement based on pure \textit{Question+Answer}~(without options), which isolates the correct knowledge without the influence of the three distractors. Details are in Appendix~\ref{sec:unc}. Meanwhile, the correctness refers to the original multiple-choice question. 

\noindent\textbf{Low Uncertainty \& Correct Answer:} LLM shows confidence about the correct knowledge and also answers the question correctly. In this case, we consider the LLM to possess robust knowledge.

\noindent\textbf{High Uncertainty \& Incorrect Answer:} LLM expresses uncertainty and answers incorrectly. So, we conclude that the LLM lacks the knowledge.

\noindent\textbf{High Uncertainty \& Correct Answer:} LLM is unsure but answers correctly. This indicates that it may retain the correct knowledge, but the memory is vague, or that the semantic shortcuts in the question lead to the correct answer.

\noindent\textbf{Low Uncertainty \& Incorrect Answer:} LLM is confident yet answers incorrectly. This could result from the LLM recalling incorrect knowledge or from misleading distractors in questions.

In the first two categories, we can determine whether the LLM truly possesses the knowledge. However, in the latter two cases, multiple factors influence the final results, and the LLM's grasp of knowledge is not robust.

This classification method leverages the strengths of both multiple-choice and generative questions, since we collect the uncertainty score without the distractors. While multiple-choice questions are easy to evaluate, they may allow for shortcuts; generative questions, on the other hand, are the opposite. Our method capitalizes on the uncertainty inherent in generative questions, which do not have shortcuts, and the accuracy of easily parsed answers provided by multiple-choice questions. This approach ensures that evaluation remains straightforward while fully addressing the potential for shortcuts.

\begin{figure*}[t]
\centering
\includegraphics[scale=0.4]{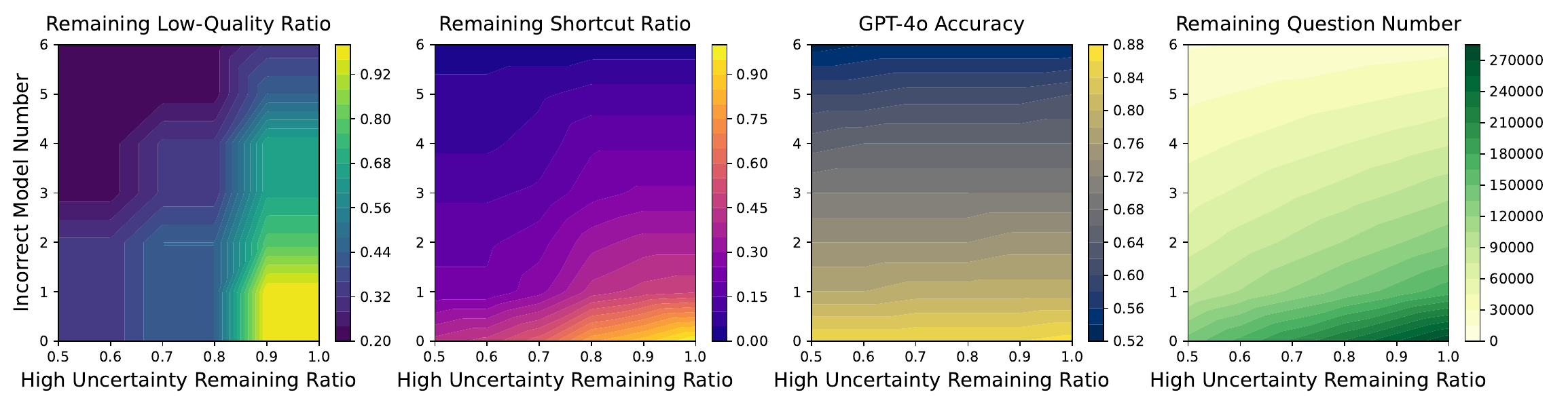}
\vspace{-0.1in}
\caption{Heatmaps illustrating how subset quality changes with \textit{incorrect model number} and \textit{high uncertainty remaining ratio}. The former refers to the minimum number of times that each remaining question in the subset is answered incorrectly. The latter refers to the proportion of high-uncertainty questions that are retained in the subset. } 
\label{filter}
\vspace{-0.15in}
\end{figure*}

\subsection{The Two-Round Measurement}

We introduce our two-round measurement, which can be applied to any multiple-choice benchmarks,  based on the four categories. In the first round, we present multiple-choice questions to evaluate LLM's performance. Then, results are classified into four categories.
For questions falling into the latter two cases~(high-uncertainty correct \& low-uncertainty incorrect), we will conduct a second round questioning. Scores will be modified if the correctness of any questions changes in this round.

For questions requiring reassessment in the second round, we ask LLMs to judge whether the four statements, with details in Appendix~\ref{sec:unc}, are true or false. The LLM is considered to truly possess the knowledge only when all four statements are accurately judged. On one hand, the second round provides an opportunity for LLMs to correct the answer by breaking distractors into separate questions, and on the other hand, it identifies questions where the LLMs simply guess the correct answers.

Compared to breaking down the multiple-choice questions into four correctness judgments directly from the start, our two-round approach offers a comprehensive analysis of how LLMs' performance changes from the first to the second stage. It leverages both uncertainty and correctness, providing deeper insights into LLMs' confidence and robustness regarding the factual knowledge they retain. Additionally, this method enables us to determine whether a particular result is due to LLM's lack of knowledge or the shortcuts and misleading distractors in the multiple-choice questions. 

Although the second round of evaluation is stricter, our \dataset{} benchmark ensures fair classification and comparison between high-frequency and low-frequency questions under the same setting. Because we have a shared abstract question for each pair of entities, controlling other factors which may affect LLM's uncertainty and accuracy. Thus, we can rule out all the other factors to purely compare the difference between high-frequency and low-frequency knowledge.

\subsection{Experiments and Analysis}
\label{sec: Robust Knowledge experiment}

Here, we conduct the experiments using this method to measure LLMs' knowledge robustness in a few-shot manner. Results are shown in Table~\ref{tab:robust_knowledge_results}

There are very interesting observations from the results: 
\textbf{(1) GPT-4o and GPT-4o-mini can not stand the test of robust knowledge measurement especially on low-frequency knowledge:} Their performance is still very good on high-frequency questions after the second round but drops a lot on low-frequency ones. The accuracy of GPT-4o-mini on low-frequency questions after the second round is only 38.25, about 35 points lower than the first round. Their average performance even drop the most among all the evaluated models. \textbf{(2) The LLMs' grasp of low-frequency knowledge is less robust than that of high-frequency knowledge:} Performance of LLMs in the second round all drop much more on low-frequency questions than on high-frequency questions. For example, in all of the open source LLMs, the accuracy of Llama-3.1-8B-Instruct in the second round, whose average performance is the best in the first round, dropped up to about 25 points on low-frequency questions while only about 7 points on high-frequency questions. \textbf{(3) LLMs which can stand the test of robust knowledge measurement must have a very accurate grasp of low-frequency knowledge:} LLMs whose average accuracy in the second round is higher than 55 are all those who performed relatively better on the low-frequency questions. This illustrates the importance of the LLMs mastering low-frequency knowledge if they want to be reliable models in the aspect of factual knowledge.

\section{\dataset-Hard}
\label{sec:ComparisonQA_Hard}

Directly collected questions, especially the high-frequency part, are a bit simple for today's LLMs. More importantly, they may have low quality and semantic shortcuts. So we introduce a new method to select a subset called \dataset-Hard, containing only difficult, low-frequency questions that have high quality and no semantic shortcuts.

\subsection{Hard High-Quality Question Filtering}
The filtering method utilizes both correctness and uncertainty, similar to the last section. 
Previous benchmarks, like SimpleQA~\cite{SimpleQA} and MMLU-Pro~\cite{mmlupro}, were collected adversarially based on LLMs' responses to ensure question difficulty. We enhance this method by also considering LLMs' uncertainty to achieve the selection of high-quality, shortcut-free questions.

As illustrated in the third part of Figure~\ref{pipeline}, we collect the correctness and uncertainty from six open-source LLMs, listed in Appendix~\ref{sec:full_Experiments}, for questions with low-frequency entities. Different from above, both metrics here are about the entire multiple-choice questions, considering all the four options' quality. We choose the questions that many models answer incorrectly and exhibit high uncertainty for our hard subset. We define two hyperparameters: \textit{incorrect model number} and \textit{high uncertainty remaining ratio}. The former is the minimum number of times that each remaining question is answered incorrectly. The latter is the proportion of high-uncertainty questions~(sort by the sum of the uncertainty of all models) retained in the subset.

The method is designed based on this assumption: Low-quality questions and those with semantic shortcuts are often associated with correct answers and low uncertainty across different models, as our benchmark is constructed by LLMs-generated questions. The rationale behind this is as follows: (1) For low-quality questions, if the question or answer is incorrect, it is likely that the LLM did not generate it from the descriptions but rather from its internal knowledge. Consequently, models may display high confidence and yield correct answers. (2) For questions containing shortcuts, models may cheat through shortcuts, resulting in lower uncertainty and higher accuracy. These will all be proved in the following experiments.

\begin{table}[t]
    \huge
    \centering
    \resizebox{\linewidth}{!}{
	\begin{tabular}{@{}l|l||c>{\columncolor{hard}}cc@{}}
	\toprule
    \textbf{Method}&\textbf{Backbone}&\textbf{Unc.}&\cellcolor{white}\textbf{Acc}&\textbf{Ma-F1}\\
            \midrule
             \multirow{12}{*}{\textbf{\begin{tabular}[c]{@{}l@{}}Open Source \\ LLM\\ \end{tabular}}}
            &Llama-3 \Large{\textit{8B}}  & 25.04 & 27.95 & 27.74 \\
            &Llama-3-Instruct \Large{\textit{8B}} & 30.57 & 22.38 & 22.32 \\
            &Llama-3.1 \Large{\textit{8B}}  & 23.87 & 28.76 & 28.40 \\
            &Llama-3.1-Instruct \Large{\textit{8B}} & 23.77 & 27.06 & 26.59 \\
            &Llama-3.2 \Large{\textit{3B}}  & 26.97 & 27.10 & 27.05 \\
            &Llama-3.2-Instruct \Large{\textit{3B}} & 27.04 & 28.68 & 28.65 \\
            &Gemma-2 \Large{\textit{2B}} & 31.07 & 22.80 & 22.35 \\
            &Gemma-2 \Large{\textit{9B}} & 23.67 & 24.71 & 24.64  \\
            &Phi-3.5-mini \Large{\textit{4B}}  & \underline{\textbf{12.42}} & \underline{31.87} & \underline{31.19} \\
            &Falcon2 \Large{\textit{11B}}  & 18.65 & 23.70 & 23.64 \\
            &Mistral-v0.3 \Large{\textit{7B}} & 15.69 & 23.96  & 23.74 \\
            &Mistral-v0.3-Instruct \Large{\textit{7B}}  & 18.36 & 23.65 & 23.25 \\
            \midrule
            \multirow{2}{*}{\textbf{\begin{tabular}[c]{@{}l@{}} Proprietary \\ LLM\\ \end{tabular}}}
            &GPT4o-mini  & \underline{46.49} & 38.13 & 37.82 \\
            &GPT4o  & 54.66 & \underline{\textbf{69.98}} & \underline{\textbf{70.02}} \\
		\bottomrule
	\end{tabular}
    }
\vspace{-0.07in}
\caption{Performance of various LLMs on the testing set of~\dataset{}-Hard. Unc., Acc, and Ma-F1, denote Uncertainty, Accuracy, and Macro F1-score. The best performances within each method are \underline{underlined} and the best among all methods are \textbf{bold-faced}.} 
\label{tab:subset_results}
\vspace{-0.07in}
\end{table}

\subsection{Parameters Chosen by Expert Verification}

To validate our filtering method, we invite experts to annotate the shortcuts in the 200 randomly sampled questions in the same setting mentioned in \S\ref{sec:Expert Verification}. 
The instruction is the same as the \textit{standard (2)} given to LLMs above.
Results show that 9.4\% of the 95.5\% correct and high quality questions are identified as having semantic shortcuts. 

Then, we examine how the following metrics change with different settings regarding correctness and uncertainty, which are illustrated in Figure~\ref{filter}: \textbf{(1) Remaining Low-Quality Ratio }(the proportion of remaining low-quality problems relative to the original number of low-quality problems), \textbf{(2) Remaining Shortcut Ratio }(the proportion of remaining problems with shortcuts relative to the original number of problems with shortcuts), \textbf{(3) GPT-4o Accuracy} (the accuracy of GPT-4o on the remaining questions), and \textbf{(4) Remaining Question Count} (the size of the remaining questions). The results suggest that both uncertainty and correctness contribute to the selection of high-quality, shortcut-free questions, thereby demonstrating the effectiveness of our method. While it is expected that different models' correctness would aid in identifying more challenging questions, it is noteworthy that uncertainty proved to be more effective in selecting high-quality and shortcut-free questions while maintaining the dataset size, validating the inclusion of uncertainty in our filtering method.

Finally, we choose to set \textit{incorrect model number} to 3 and \textit{high uncertainty remaining ratio} to 0.8 according to Figure~\ref{filter}. It is a trade-off between quality, difficulty, and subset size. In this setting, only 1.5\% of the total questions are of low quality, and 2.1\% with shortcuts. Finally, the subset size is 81K, with a GPT-4o accuracy of 70\%.
Detailed statistics are shown in Appendix~\ref{sec:statistics}.

\subsection{Experiments and Analysis}

Then we conduct experiments on \dataset-Hard in a few-shot manner, with results shown in Table~\ref{tab:subset_results}. In the multiple-choice question format, the open-source LLMs all significantly underperform, indicating the difficulty of our benchmark. For the proprietary LLMs, GPT-4o-mini also has a poor performance with an accuracy of about 38, even though we do not use it when constructing the subset. GPT-4o is better with an accuracy of about 70, but still has a huge room for future enhancement. And, predictably, its knowledge robustness will drop much more on this benchmark according to our experiments in \S\ref{sec: Robust Knowledge experiment}.

\section{Conclusions}

In this paper, we first introduce \dataset{} benchmark to evaluate LLMs' factual knowledge, with a fully automatic pipeline. This benchmark allows for more controllable and detailed comparisons between high-frequency and low-frequency knowledge of LLMs. Then, we propose a two-round method utilizing correctness and uncertainty to measure LLMs' knowledge robustness. And we are surprised to find that even powerful LLMs like GPT-4o can not stand such a test, especially on the low-frequency knowledge. At last, we discover that uncertainty is more effective in filtering out questions with low quality and shortcuts compared with correctness, which is often used by recent works. Based on this method, we provide a subset called \dataset{}-Hard, which contains only difficult low-frequency questions of high quality and no shortcuts for future study.

\clearpage
\section*{Limitations}

While we contribute valuable resources, methods, and findings to advance the probing of LLMs' factual knowledge, several limitations still exist that cannot be covered in this single work. 

In this paper, we provide a shared abstract question with the entities being the only varying part. However, due to the sharing of such abstract questions, it is difficult to contain more entities with different frequencies in the same question because many entities lack sufficient shared features to generate common questions.

Our approach ensures that the difference between a pair is only the entity. Future research could investigate methods for measuring the entire knowledge frequency required to solve the questions, rather than limiting it to entity frequency. We believe it is a challenging but valuable task.


Additionally, our focus is on fixed knowledge, but fast-changing factual knowledge~\cite{do2024really} and other knowledge~\cite{do2024constraintchecker, wang2022subeventwriter, wang2023cola} also deserves attention. Specifically, exploring how knowledge frequency can assist LLMs in acquiring new information~\cite{choi2023kcts, zong2023tilfa} is a worthwhile area for further study.

\section*{Ethics Statement}
\noindent\textbf{Offensive Content Elimination.}
Our benchmark curation pipeline, which involves generating content using LLMs, requires stringent measures to ensure that generated responses are free from offensive material. 
We manually review a random sample of 200 data instances from \dataset{} for any offensive content. 
Based on our annotations, we have not detected any offensive content.
Therefore, we believe our dataset is safe and will not yield any negative societal impact.

\noindent\textbf{Licenses.}
We will share our code under the MIT license, allowing other researchers free access to our resources for research purposes. 
Our dataset will be released under a CC license, also providing scholars with free access.
We take full responsibility for any rights violations or issues related to the data license.
The DBpedia dataset used in this paper is shared under the CC BY-SA license, permitting its use for research. 
As for language models, we access all open-source LMs via the Huggingface Hub~\cite{DBLP:conf/emnlp/WolfDSCDMCRLFDS20}. 
All associated licenses permit user access for research purposes, and we have agreed to adhere to all terms of use.

\noindent\textbf{Annotations.}
For expert verifications, we have obtained IRB approval and support from our institution's department, enabling us to invite expert graduate students to validate the quality of our data. 
They all agree to participate voluntarily without being compensated.
We have made significant efforts to eliminate offensive content, thereby ensuring that no annotators are offended.

\section*{Acknowledgments}

The authors of this paper were supported by the ITSP Platform Research Project (ITS/189/23FP) from ITC of Hong Kong, SAR, China, and the AoE (AoE/E-601/24-N), the RIF (R6021-20) and the GRF (16205322) from RGC of Hong Kong,SAR, China.

\bibliography{custom}

\newpage
\appendix
\begin{center}
    {\Large\textbf{Appendices}}
\end{center}

\section{Benchmark Statistics}
\label{sec:statistics}

\dataset{} is a large-scale benchmark comprising a total of 283,455 abstract question pairs, each paired with a high-frequency and a low-frequency entity. 
We guarantee that each entity corresponds to a single question sourced from 9166 hypernyms, to ensure no overlap.
We partition our data into training, validation, and testing splits following an 8:1:1 ratio, ensuring that entities of different frequency intervals are evenly distributed in each split. 
Specific details are shown in Table~\ref{dataset_scale}.

And for \dataset-Hard, there are 81,136 high-quality and shortcut-free questions with low-frequency entities from 4,876 hypernyms in total, with details shown in Table~\ref{dataset_scale_hard}.

\begin{table}[h]
    \large
	\centering
    \resizebox{\linewidth}{!}{
	\begin{tabular}{c|ccccc}
	\toprule
        \textbf{Split} & \textbf{\#Q. Pair} & \textbf{\#Entity} & \textbf{\#Hyper.} & \textbf{\#H. R.} & \textbf{\#L. R.}\\
        \midrule
        Train & 226,762 & 453,524 & 8,430 & 628 & 76\\
        Valid & 28,345 & 56,690 & 3,316 & 627 & 76\\
        Test & 28,348 & 56,696 & 3,314 & 629 & 76\\
        \midrule
        Total & 283,455 & 566,910 & 9,166 & 628 & 76\\
        \bottomrule
        \end{tabular}
        }
        \caption{The statistics of \dataset{} benchmark. \#Q. Pair refers to the number of question pairs. Hyper. means hypernym. \#H. R. and \#L. R. refer to the average number of relationships of high-frequency entities and low-frequency entities respectively.}
    \label{dataset_scale}
\end{table}

\begin{table}[h]
    \small
	\centering
	\begin{tabular}{c|cccc}
	\toprule
        \textbf{Split} & \textbf{\#Q. = \#Entity} & \textbf{\#Hyper.}  & \textbf{\#L. R.}\\
        \midrule
        Train & 65,057 & 4,396 & 77.45\\
        Valid & 7,978 & 1,466 & 77.23\\
        Test & 8,101 & 1,484 & 77.65\\
        \midrule
        Total & 81,136 & 4,876 & 77.44\\
        \bottomrule
        \end{tabular}
        \caption{The statistics of \dataset{}-Hard benchmark. Here the number of questions is equal to the number of entities since it only has low frequency entities.}
    \label{dataset_scale_hard}
\end{table}

For benchmark quality, we further compute the correctness of the annotated date in \S\ref{sec:Expert Verification} coming from the train, validation, and test set separately. The results presented in Table~\ref{dataset_quality} show the high correctness in all splits of our benchmark. Besides, our labeling is very strict. If either of the questions in one pair is wrong, we will judge the whole question pair incorrect. Since each pair contains two questions, actually only about 2.25\% of the total questions are incorrect.

\begin{table}[h]
    \small
	\centering
	\begin{tabular}{c|ccc}
	\toprule
        \textbf{Split} & \textbf{Correctness} & \textbf{Agreement}  \\
        \midrule
        Train & 95.68 & 88.27 \\
        Valid & 93.75 & 87.50 \\
        Test & 95.45 & 90.91\\
        \bottomrule
        \end{tabular}
        \caption{The statistics of \dataset{}-Hard benchmark. Here the number of questions is equal to the number of entities since it only has low frequency entities.}
    \label{dataset_quality}
\end{table}

\begin{table*}[h]
        \small
	\centering
	\begin{tabularx}{\textwidth}{l|X}
		\toprule
        Task&Prompt \\
        \midrule
	
        Question Generation & You will be given two entities belonging to the same hypernym. Generate a shared multiple choice question for both entities based on their descriptions according to the following 5 steps.
        
Requirements: First, the questions should have one and only one correct answer for both entities. Second, the correct answer cannot be simply guessed by the names of the entities or the way the questions are asked.
For example, the question: "What was the primary operational location of Sydney O-Class Tram? A. Oslo, Norway B. Sydney, Australia C. Stockholm, Sweden D. Melbourne, Australia" is not allowed since only the correct answer contains "Sydney" which is also in the entity's name.
Third, the four options should have roughly the same length.

Step 1, generate the shared question containing "[entity\_name]" which can be replaced by the two entity names and then have different answers accordingly.

Step 2, use roughly the same amount of words to answer the questions for both entities separately.

Step 3, generate a misleading distractor for both entities separately. The distractors should have roughly the same length with the correct answers.

Step 4, form the final multiple choice question using the above four answer candidates, and make sure the answer for Entity1 is 'A' , the answer for Entity2 is 'B', and their misleading answer candidates are 'C' and 'D'.

Step 5, check whether the final question and answer candidates meet the above requirements. If yes, then output **SCUUEED**, otherwise output **FAIL**.

\\
&Follow these examples: 

……\textit{(Examples written by experts)}











\\
&\textbf{Hypernym:} \textit{[Hypernym]}

\textbf{Entity1:} \textit{[Entity1]}

\textbf{Entity1 Description:} \textit{[Entity1 Description]}

\textbf{Entity2:} \textit{[Entity2]}

\textbf{Entity2 Description:} \textit{[Entity2 Description]}
\\
        \bottomrule
	\end{tabularx}
     \vspace{-0.1in}
	\caption{The prompt used to generate questions in \dataset. Placeholders \textit{[Hypernym]}, \textit{[Entity1]}, \textit{[Entity1 Description]}, \textit{[Entity2]}, \textit{[Entity2 Description]} will be replaced with real hypernym, high-frequency entity, low-frequency entity, and their descriptions accordingly.}
	\label{gen_prompt}
 \vspace{-0.1in}
\end{table*}

\section{Definition of High and Low Frequency}
\label{sec:freq}

First, we randomly sample 1K entities from DBpedia and compute their relationship count separately. 
This process is to study the distribution of entities’ relationship counts in DBpedia and find boundaries for high-frequency and low-frequency entities. 
Then we sort these entities in order of their relationship count from highest to lowest. 
High-frequency entities possess cumulative relationship count of up to 1/3 of all entities, while low frequency entities range from 2/3 to 1. We exclude those between 1/3 and 2/3 to make comparison clear.

Repeating this process 3 times, we get the number of relationships to distinguish high-frequency and low-frequency entities, which are higher than 185 and lower than 107 respectively. (Of the 1K randomly sampled entities, 119 entities are identified as high frequent and 621 entities as low frequent.) These two numbers are then used to classify all the high-frequency and low-frequency entities. 

It is necessary to randomly select 1K entities first to calculate the two boundaries, since the time to compute the cumulative relationship count for all the sorted entities in DBpedia is unaffordable.

\section{Details in \dataset{} Construction}
\label{sec:generatin}
In the process of entity pairs extraction, both hypernyms that do not have enough entities and low-frequency entities that do not have enough high-frequency entities to pair with are all discarded. Finally, we get 293K entities pairs from 9,261 hypernyms in total. 

These pairs are then fed into the LLM to generate questions. Each pair has a shared abstract question respectively. Since a hypernym can have several entity pairs, the final number of abstract questions, which is equal to the number of pairs, is more than that of hypernyms. After LLM's proofread stage, there are 283K abstract questions left, which are used to build our \dataset{} benchmark. Entities belonging to these questions come from 9,166 different hypernyms.

In the process of Abstract Question Generation, the prompt we use to generate questions is shown in Table~\ref{gen_prompt}.
In questions generated by LLMs, the answer for the high-frequency entity is $A$, low-frequency entity is $B$, and their distractors are $C$ and $D$ respectively. And in the end, we randomly shuffle the 4 options in one question to guarantee the balance of the correct options.

\begin{table}[h]
        \small
	\centering
	\begin{tabular}{p{1.4cm}|p{5.3cm}}
		\toprule
        Task&Prompt \\
        \midrule
	
        Uncertainty& Question: [question]\\
Generation& Answer: **[option].** \\
&Uncertainty: **[uncertainty percentage]** \\
&Answer the following multiple choice question. Select only one correct answer from the choices and give your uncertainty score, following the above format.\\\\
&\textit{[Question]}  \\
& A. \textit{[OptionA]}.  B. \textit{[OptionB]}.\\
& C. \textit{[OptionC]}.  D. \textit{[OptionD]}.\\
        \bottomrule
	\end{tabular}
     \vspace{-0.1in}
	\caption{The prompt used to generate uncertainty score for proprietary LLMs. Placeholders \textit{[Hypernym]}, \textit{[Entity1]}, \textit{[Entity1 Description]}, \textit{[Entity2]}, \textit{[Entity2 Description]} will be replaced with real hypernym, high-frequency entity, low-frequency entity, and their descriptions accordingly.}
	\label{unc_prompt}
 \vspace{-0.2in}
\end{table}

\section{Calculation of Uncertainty}
\label{sec:unc}

Following \citet{vera}, we first combine the questions with each of their options, and use GPT-4o-mini to transform them into four statements.  The data can also be found in our benchmark. 

Then we compute the uncertainty for each statement.
For open-source LLMs, uncertainty refers to their perplexity for generating the correct statement. 
For proprietary LLMs, we allow them to generate their uncertainty scores, with prompt shown in Table~\ref{unc_prompt}. 
Then, the threshold between high and low uncertainty for each model is determined by its own average uncertainty on the testing set of each benchmark respectively.

There are several reasons for choosing different uncertainties for each setting. On one hand, perplexity-based uncertainty is unavailable for proprietary LLMs. On the other hand, open source LLMs often fail to understand the instructions and can not generate uncertainty scores. We have conducted experiments on verbalized uncertainty of open source LLMs. However, experiments show that Llama-3-8B-Instruct has a 18.5\% chance of not generating an uncertainty score, which makes the results unreliable and will also affect further robustness evaluations. In addition, \citet{UQ} mentioned that, for multiple choice QA, information-based methods such as perplexity are substantially superior to quantifying model uncertainty.

\begin{table}[h]
        \small
	\centering
	\begin{tabular}{p{1.3cm}|p{5.5cm}}
		\toprule
        Method&Prompt \\
        \midrule
	Zero-Shot &\textit{[Question]}  \\
    &A. \textit{[OptionA]}.  B. \textit{[OptionB]}.  \\
    &C. \textit{[OptionC]}.  D. \textit{[OptionD]}.
    \\ & The correct answer is: \\
    \midrule
        Few-Shot & \textit{[Examples]} \\
&Answer the multiple choice question. Select only one correct answer from the choices, following above examples. \\\\
&\textit{[Question]}  \\
& A. \textit{[OptionA]}.  B. \textit{[OptionB]}.\\
& C. \textit{[OptionC]}.  D. \textit{[OptionD]}.\\
\midrule
        CoT & Question: [question]\\
& Rational: [rationale]\\
&Answer: **[option].** \\
&Answer the multiple choice question. Think step by step and generate a short rationale to support your reasoning. Choose one best answer based on the generated rational, following the above format. Keep your whole response in 50 tokens. \\\\
&\textit{[Question]}  \\
& A. \textit{[OptionA]}.  B. \textit{[OptionB]}.\\
& C. \textit{[OptionC]}.  D. \textit{[OptionD]}.\\
        \bottomrule
	\end{tabular}
     \vspace{-0.1in}
	\caption{The prompt used when evaluating LLMs on our benchmark. Placeholders \textit{[Examples]}, \textit{[Question]}, \textit{[OptionA]}, \textit{[OptionB]}, \textit{[OptionC]}, \textit{[OptionD]} will be replaced with the real examples, questions and their options accordingly.}
	\label{eval_prompt}
 \vspace{-0.15in}
\end{table}

\section{Experiment Details}
\label{sec:full_Experiments}

\begin{table*}[h]
    \huge
    \centering
    \resizebox{\linewidth}{!}{
	\begin{tabular}{@{}l||c>{\columncolor{high}}ccc>{\columncolor{low}}cc|c>{\columncolor{total}}cc||ccc@{}}
	\toprule
    \multirow{3}{*}{\textbf{Models}}&\multicolumn{3}{c}{\textbf{High Freq Question}} &\multicolumn{3}{c|}{\textbf{Low Freq Question}}&\multicolumn{3}{c|}{\textbf{Average}}&\multicolumn{3}{c}{\textbf{Difference (H --> T)}}
    \\
	&\textbf{Unc.}&\cellcolor{white}\textbf{Acc}&\textbf{Ma-F1}&\textbf{Unc.}&\cellcolor{white}\textbf{Acc}&\textbf{Ma-F1}&\textbf{Unc.}&\cellcolor{white}\textbf{Acc}&\textbf{Ma-F1}&\textbf{Unc.}&\cellcolor{white}\textbf{Acc}&\textbf{Ma-F1}\\
    &\textbf{(↓)}&\cellcolor{white}\textbf{(↑)}&\textbf{(↑)}&\textbf{(↓)}&\cellcolor{white}\textbf{(↑)}&\textbf{(↑)}&\textbf{(↓)}&\cellcolor{white}\textbf{(↑)}&\textbf{(↑)}&&&\\
            \midrule
            Random & - & 25.29 & 25.29 & - & 25.22 & 25.22 & - & 25.26 & 25.26 & - & ↓ 0.07 & ↓ 0.07 \\
            Majority & - & 25.70 & 10.22 & - & 25.14 & 10.04 & - & 25.42 & 10.13 & - & ↓ 0.56 & ↓ 0.18 \\
		  \bottomrule
          \rowcolor[gray]{0.9} \multicolumn{13}{c}{\textbf{LLM (Open Source) + Zero-shot}} \\
          \toprule
            Llama-3 \Large{\textit{8B}} & 54.33 & 65.90 & 63.60 & 81.54 & 53.83 & 51.29 & 67.94 & 59.87  & 57.44  & ↑ 6.11 & ↓ 12.07 & ↓ 12.30 \\
            Llama-3-Instr \Large{\textit{8B}} & 77.55 & \underline{80.72} & \underline{80.71} & 117.41 & 69.03 & 68.95 & 97.48  & 74.88 & 74.83 & ↑ 39.86 & ↓ 11.69 & ↓ 11.76 \\
            Llama-3.1 \Large{\textit{8B}} & 55.91 & 65.29 & 63.31 & 83.35 & 52.66 & 50.52 & 69.63  & 58.98  & 56.92  & ↑ 27.44 & ↓ 12.63 & ↓ 12.78 \\
            Llama-3.1-Instr \Large{\textit{8B}} & 58.97 & 80.06 & 80.08 & 87.05 & \underline{69.99} & \underline{69.94} & 73.01  & \underline{75.03}  & \underline{75.01}  & ↑ 28.08 & ↓ 10.07 & ↓ 10.14 \\
            Llama-3.2 \Large{\textit{3B}} & 67.51 & 57.93 & 53.65 & 99.55 & 48.10 & 44.66 & 83.53  & 53.02  & 49.16  & ↑ 32.04 & ↓ 9.83 & ↓ 9.00\\
            Llama-3.2-Instr \Large{\textit{3B}} & 74.27 & 71.47 & 71.51 & 108.24 & 62.17 & 62.18 & 91.26  & 66.82  & 66.85  & ↑ 33.97 & ↓ 9.30 & ↓ 9.33 \\
            Gemma-2 \Large{\textit{2B}} & 133.57 & 46.48 & 43.40 & 213.40 & 37.76 & 34.07 & 173.49  & 42.12  & 38.74 & ↑ 79.83  & ↓ 8.72 & ↓ 9.33 \\
            Gemma-2 \Large{\textit{9B}} & 124.97 & 64.50 & 64.80 & 211.34 & 52.24 & 51.23 & 168.16  & 58.37  & 58.01  & ↑ \underline{\textbf{86.37}} & ↓ 12.26 & ↓ \underline{13.57} \\
            Phi-3.5-mini-Instr \Large{\textit{4B}} & \underline{27.81} & 72.81 & 72.78 & \underline{39.80} & 65.26 & 65.00 & \underline{33.81}  & 69.04  & 68.89  & ↑ 11.99 & ↓ 7.55 & ↓ 7.78 \\
            Falcon \Large{\textit{7B}} & 62.33 & 18.19 & 18.55 & 92.92 & 17.93 & 18.19 & 77.63  & 18.06  & 18.37  & ↑ 30.59 & ↓ 0.26 & ↓ 0.36 \\
            Falcon-Instr \Large{\textit{7B}} & 88.88 & 25.68 & 14.49 & 128.75 & 25.71 & 14.01 & 108.82  & 25.70  & 14.25  & ↑ 39.87 & ↑ 0.02 & ↓ 0.48\\
            Falcon2 \Large{\textit{11B}} & 56.93 & 70.72 & 69.80 & 87.31 & 58.07 & 56.70 & 72.12  & 64.40  &  63.25 & ↑ 30.38 & ↓ \underline{12.65} & ↓ 13.10 \\
            Mistral-v0.3 \Large{\textit{7B}} & 39.83 & 65.55 & 63.11 & 56.99 & 53.36 & 50.26 & 48.41  & 59.46  & 56.69  & ↑ 17.16 & ↓ 12.19 & ↓ 12.85 \\
            Mistral-v0.3-Instr \Large{\textit{7B}} & 44.73 & 73.53 & 73.14 & 66.09 & 63.05 & 62.40 & 55.41  & 68.29  & 67.77  & ↑ 21.36 & ↓ 10.48 & ↓ 10.74\\
            \bottomrule
          \rowcolor[gray]{0.9} \multicolumn{13}{c}{\textbf{LLM (Open Source) + 4-shot}} \\
          \toprule
            Llama-3 \Large{\textit{8B}} & 21.89 & 75.57 & 75.55 & 23.75 & 61.00 & 61.01 & 22.82  & 68.29  & 68.28  & ↑ 1.86 & ↓ \underline{\textbf{14.57}} & ↑ \underline{\textbf{14.55}} \\
            Llama-3-Instr \Large{\textit{8B}} & 26.20 & 79.94 & 79.92 & 28.70 & 67.98 & 67.95 & 27.45  & 73.96  & 73.93  & ↑ \underline{2.50} & ↓ 11.96 & ↓ 11.96 \\
            Llama-3.1 \Large{\textit{8B}} & 20.82 & 74.91 & 74.89 & 22.62 & 62.00 & 62.00 & 21.72  & 68.46  & 68.45  & ↑ 1.81 & ↓ 12.91 & ↓ 12.90 \\
            Llama-3.1-Instr \Large{\textit{8B}} & 20.63 & 79.74 & 79.74 & 22.48 & \underline{69.09} & \underline{69.07} & 21.56 & \underline{74.42}  & \underline{74.40}  & ↑ 1.85 & ↓ 10.65 & ↓ 10.66 \\
            Llama-3.2 \Large{\textit{3B}} & 23.58 & 68.89 & 68.84 & 25.59 & 57.26 & 57.17 & 24.59  & 63.08  & 63.01  & ↑ 2.01 & ↓ 11.63 & ↓ 11.67 \\
            Llama-3.2-Instr \Large{\textit{3B}} & 23.66 & 71.43 & 71.43 & 25.67 & 62.12 & 62.09 & 24.67  & 66.78  & 66.76  & ↑ 2.01 & ↓ 9.32 & ↓ 9.34 \\
            Gemma-2 \Large{\textit{2B}} & 26.96 & 62.99 & 62.93 & 29.43 & 50.93 & 50.91 & 28.20  & 56.96  & 56.92  & ↑ 2.47 & ↓ 12.06 & ↓ 12.02 \\
            Gemma-2 \Large{\textit{9B}} & 20.26 & \underline{80.10} & \underline{80.08} & 22.36 & 68.36 & 68.31 & 21.31  & 74.23  & 74.20  & ↑ 2.10 & ↓ 11.74 & ↓ 11.77 \\
            Phi-3.5-mini-Instr \Large{\textit{4B}} & \underline{11.02} & 73.68 & 73.67 & \underline{\textbf{11.85}} & 67.46 & 67.33 & \underline{11.44}  & 70.57  & 70.50  & ↑ 0.83 & ↓ 6.22 & ↓ 6.34 \\
            Falcon \Large{\textit{7B}} & 21.69 & 28.40 & 21.25 & 23.42 & 28.25 & 20.63 & 22.56  & 28.33  & 20.94  & ↑ 1.73 & ↓ 0.15 & ↓ 0.62\\
            Falcon-Instr \Large{\textit{7B}} & 21.69 & 26.27 & 18.32 & 23.42 & 25.67 & 18.15 & 22.56  & 25.97  & 18.23  & ↑ 1.73 & ↓ 0.59 & ↓ 0.17 \\
            Falcon2 \Large{\textit{11B}} & 16.42 & 77.11 & 77.01 & 17.77 & 65.92 & 65.75 & 17.10  & 71.52  &  71.38 & ↑ 1.35 & ↓ 11.19 & ↓ 11.26 \\
            Mistral-v0.3 \Large{\textit{7B}} & 13.89 & 75.55 & 75.53 & 14.99 & 62.88 & 62.85 & 14.44  & 69.22  & 69.19   & ↑ 1.10 & ↓ 12.67 & ↓ 12.68\\
            Mistral-v0.3-Instr \Large{\textit{7B}} & 15.97 & 74.46 & 74.40 & 17.40 & 65.51 & 65.35 & 16.69  & 69.99  & 69.87  & ↑ 1.43 & ↓ 8.95 & ↓ 9.05\\
            \bottomrule
          \rowcolor[gray]{0.9} \multicolumn{13}{c}{\textbf{LLM (Proprietary) API}} \\
          \toprule
            GPT4o-mini (Zero-Shot) & 13.52 & 85.61 & 85.58 & 18.34 & 73.85 & 73.73 & 15.93  & 79.73  & 79.66  & ↑ 4.82 & ↓ 11.76 & ↓ 11.85 \\
            GPT4o-mini (Few-Shot) & 25.74 & 84.78 & 84.69 & 38.17 & 72.76 & 72.47 & 31.96  & 78.77  & 78.58  & ↑ 12.43 & ↓ \underline{12.02} & ↓ \underline{12.22} \\
            GPT4o-mini (CoT) & 10.53 & 86.25 & 86.25 & \underline{12.27} & 74.39 & 74.40 & \underline{\textbf{11.40}} & 80.32 & 80.32 & ↑ 1.74 & ↓ 11.85 & ↓ 11.85\\
            GPT4o (Zero-Shot) & 14.18 & \underline{\textbf{93.86}} & \underline{\textbf{93.95}} & 30.98 & 85.76 & 86.69 & 22.58  & 89.81  & 90.32  & ↑ 16.80 & ↓ 8.10 & ↓ 7.26\\
            GPT4o (Few-Shot) & 28.41 & 93.94 & \underline{\textbf{93.95}} & 45.81 & \underline{\textbf{86.54}} & \underline{\textbf{86.75}} & 37.11  & \underline{\textbf{90.24}}  & \underline{\textbf{90.35}}  & ↑ \underline{17.40} & ↓ 7.40 & ↓ 7.20\\
            GPT4o (CoT) & \underline{\textbf{10.39}} & 92.40 & 92.47 & 18.36 & 85.47 & 85.72 & 14.38 & 88.93 & 89.10 & ↑ 7.97 & ↓ 6.93 & ↓ 6.75\\
		\bottomrule
	\end{tabular}
    }
\vspace{-0.07in}
\caption{Performance of various LLMs on the testing set of~\dataset{}. Unc., Acc, and Ma-F1, denote Uncertainty, Accuracy, and Macro F1-score. And the Difference column shows how scores change from high-frequency questions to low-frequency questions. The best performances within each method are \underline{underlined} and the best among all methods are \textbf{bold-faced}. And for the Difference column, We \underline{underline} the largest difference within each method and \textbf{bold} the one among all methods.} 
\label{tab:main_evaluation_results_all}
\vspace{-0.07in}
\end{table*}

For the main evaluations on \dataset{}, we categorize the evaluation of different models into three types: \textbf{(1) \textsc{Open Source LLM Zero-Shot}:} We first evaluate Llama3, Llama3.1, Llama3.2~\cite{LLAMA2,LLAMA3}, Gemma2~\cite{Gemma,Gemma2}, Phi3.5~\cite{Phi35}, Falcon, Falcon2~\cite{Falcon2}, Mistral~\cite{Mistral}, and their instruction versions accordingly in a zero-shot manner~\cite{DBLP:conf/emnlp/QinZ0CYY23}. 
\textbf{(2) \textsc{Open Source LLM Few-Shot}:} Then we evaluate the above models in a few-shot manner~\cite{FewShot}. Since our benchmark is in the form of four options multiple-choice questions, the shot number is set to four to minimize bias, where each of the four examples corresponds to a different correct answer in (a, b, c, d).
\textbf{(3) \textsc{Proprietary LLM API}:} Finally, we evaluate the performance of GPT-4o~\cite{GPT4,GPT4o} and GPT-4o-mini~\cite{GPT4omini}, using zero-shot, few-shot, and Chain-of-Thought (CoT;~\citealp{COT}). 

All the open-source models are run on 4 NVIDIA A6000 (40G) GPUs with BF32. And for proprietary LLM, we access them via OpenAI API \footnote{\url{https://platform.openai.com/docs/api-reference}}.
The different kinds of prompts we use are shown in Table~\ref{eval_prompt}.
And all the evaluation results are reported in Table~\ref{tab:main_evaluation_results_all}.

For the question filtering of \dataset-Hard, the six open-source LLMs we use are Llama-3-8B-Instruct, Llama-3.1-8B-Instruct, gemma-2-9b, Phi-3.5-mini-instruct, Falcon-11B, and Mistral-7B-Instruct-v0.3.

\end{document}